\pdfoutput=1

\documentclass[11pt]{article}

\usepackage[]{EMNLP2022}

\usepackage{times}
\usepackage{latexsym}
\usepackage[normalem]{ulem}

\usepackage{xcolor}
\usepackage{pgfplots}
\pgfplotsset{compat=1.18}
\usepackage{multirow}

\usepackage{mathtools}
\usepackage{todonotes}
\usepackage{hyperref}
\usepackage{tcolorbox}
\usepackage[capitalise]{cleveref}

\usepackage{algorithm}
\usepackage{amsmath}
\usepackage{algorithmicx}
\usepackage{algpseudocode}
\usepackage{amsfonts}
\usepackage{amssymb}
\usepackage{tabularx}
\makeatletter
\newcommand*\iftodonotes{\if@todonotes@disabled\expandafter\@secondoftwo\else\expandafter\@firstoftwo\fi}  %
\makeatother

\definecolor{mygreen}{RGB}{0, 153, 0}
\definecolor{myblue}{RGB}{51, 51, 255}
\definecolor{myorange}{RGB}{255, 128, 0}
\newcommand\errordescription[0]{\textcolor{myblue}{Error Description}}

\newcommand\nnfootnote[1]{%
  \begin{NoHyper}
  \renewcommand\thefootnote{}\footnote{#1}%
  \addtocounter{footnote}{-1}%
  \end{NoHyper}
}

\usepackage[T1]{fontenc}

\usepackage[utf8]{inputenc}

\usepackage{microtype}

\usepackage{inconsolata}

\newcommand{\vparam}{\vtheta}

\newcommand\cut[1]{}

\newcommand{\squishlist}{
   \begin{list}{$\bullet$}
    { \setlength{\itemsep}{0pt}      \setlength{\parsep}{3pt}
      \setlength{\topsep}{3pt}       \setlength{\partopsep}{0pt}
      \setlength{\leftmargin}{1.5em} \setlength{\labelwidth}{1em}
      \setlength{\labelsep}{0.5em} } }

\newcommand{\squishlisttwo}{
   \begin{list}{$\bullet$}
    { \setlength{\itemsep}{0pt}    \setlength{\parsep}{0pt}
      \setlength{\topsep}{0pt}     \setlength{\partopsep}{0pt}
      \setlength{\leftmargin}{2em} \setlength{\labelwidth}{1.5em}
      \setlength{\labelsep}{0.5em} } }

\newcommand{\squishend}{
    \end{list}  }

{}
{}
{}
{}

\newcommand{\myvec}[1]{\mbox{$\mathbf{#1}$}}
\newcommand{\myvecsym}[1]{\mbox{$\boldsymbol{#1}$}}

\newcommand{\vtheta}{\mbox{$\myvecsym{\theta}$}}

\newcommand{\va}{\mbox{$\myvec{a}$}}

\newcommand{\ve}{\mbox{$\myvec{e}$}}

\newcommand{\vk}{\mbox{$\myvec{k}$}}

\newcommand{\vq}{\mbox{$\myvec{q}$}}

\newcommand{\vs}{\mbox{$\myvec{s}$}}

\newcommand{\vu}{\mbox{$\myvec{u}$}}
\newcommand{\vv}{\mbox{$\myvec{v}$}}

\newcommand{\be}{\begin{equation}}
\newcommand{\ee}{\end{equation}}
\newcommand{\bea}{\begin{eqnarray}}
\newcommand{\eea}{\end{eqnarray}}
\newcommand{\beaa}{\begin{eqnarray*}}
\newcommand{\eeaa}{\end{eqnarray*}}

\newcommand{\blockcomment}[1]{}

\makeatother

\newcommand{\studentsol}{\vs_{\text{\vq}}}
\newcommand{\correctsol}{\widehat{\vs}_{\text{\vq}}}

\title{Stepwise Verification and Remediation of Student Reasoning Errors\\ with Large Language Model Tutors}

\author{
    Nico Daheim$^{\ast 1}$ \quad
    Jakub Macina$^{\ast 2, 3}$ \quad \\
    \textbf{
    Manu Kapur$^4$ \quad
    Iryna Gurevych$^1$ \quad
    Mrinmaya Sachan$^{2}$
    } \\ \text{} \\
  $^{1}$ Ubiquitous Knowledge Processing Lab (UKP Lab), Department of Computer Science\\ and Hessian Center for AI (hessian.AI), TU Darmstadt \\
  $^{2}$Department of Computer Science, ETH Zurich \quad
  $^3$ ETH AI Center \\
  $^4$Professorship for Learning Sciences and Higher Education, ETH Zurich \\
}

\begin{document}
\maketitle
\nnfootnote{$^\ast$Equal contribution.}
\begin{abstract}
Large language models (LLMs) present an opportunity to scale high-quality personalized education to all. 
A promising approach towards this means is to build dialog tutoring models that scaffold students' problem-solving. 
However, even though existing LLMs perform well in solving reasoning questions, they struggle to precisely detect student's errors %
and tailor their feedback to these errors. 
Inspired by real-world teaching practice where teachers identify student errors and customize their response based on them,
we focus on verifying student solutions and show how grounding to such verification improves the overall quality of tutor response generation. 
We collect a dataset of 1K stepwise math reasoning chains with the first error step annotated by teachers. We show empirically that finding the mistake in a student solution is challenging for current models. We propose and evaluate several verifiers for detecting these errors.
Using both automatic and human evaluation we show that the student solution verifiers steer the generation model towards highly targeted responses to student errors which are more often correct with less hallucinations compared to existing baselines.
\end{abstract}
\hspace{0em}\includegraphics[width=0.9em,height=0.9em]{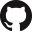}\hspace{.25em}\parbox{\dimexpr\linewidth-2\fboxsep-2\fboxrule}{\url{https://github.com/eth-lre/verify-then-generate}}
\vspace{-.5em}

\section{Introduction}
\begin{figure*}[t!]%
	\centering
	\resizebox{\textwidth}{!}{\includegraphics{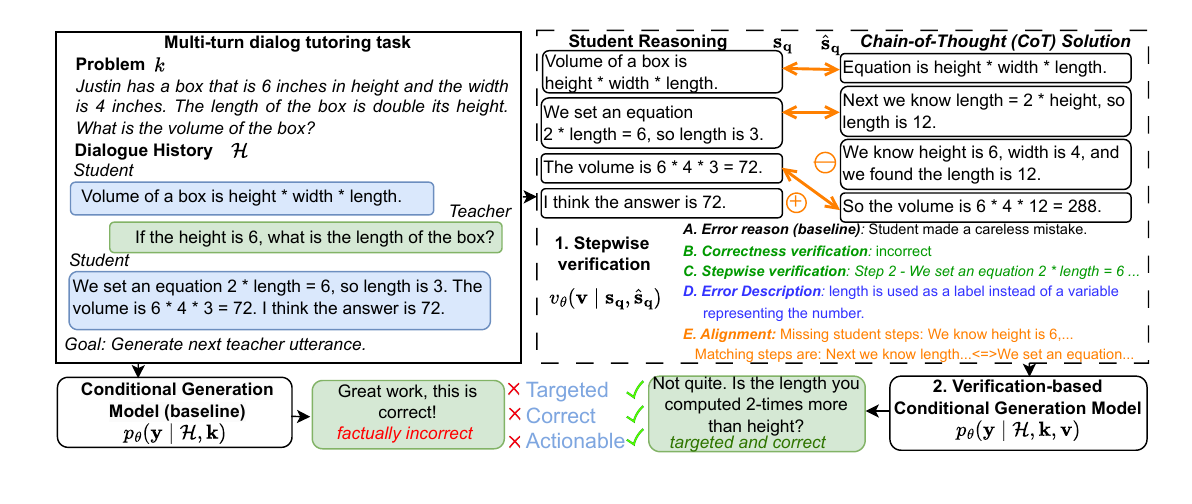}}
        \vspace{-12mm}
	\caption{Directly generating a tutor response based on the conversation history can lead to hallucinations (bottom left). To alleviate this, we split this process into two sequential tasks (right): 1) A model identifies the student's mistake. 2) A different response generation model communicates the identified mistake. 
    We use different verifiers: providing the {\bf Error Reason}~\citep{wang-etal-2024-bridging}, {\color{mygreen} Classification-based Verification}, providing a more detailed {\errordescription} and a {\color{myorange} Step Alignment} of student and reference solution.
    Especially the latter two reduce hallucinations and make tutor models more targeted at the student error when verification and generation are combined (\cref{sec:results}).
 }\label{fig:overview}
\end{figure*}
The field of dialog tutoring aims to build systems that can teach students by holding a conversation with them~\citep{wollny2021we, learnlm2024}.
Dialog tutors hold the potential to make personalized teaching available to learners anywhere anytime. The increasing capabilities of LLMs have brought renewed hope to this field ~\citep{thoppilan2022lamda, learnlm2024}.
However, real-time tutoring is quite complex, and human teachers bring various intricate capabilities when teaching, such as identifying student errors in problem solving, picking a pedagogical strategy, and communicating it~\citep{wang-etal-2024-bridging}.
The same requirements hold for dialog tutoring models which need all these abilities to be effective.

Yet, although research on effective human tutors shows they perform these steps sequentially by first reasoning about the error, then picking a strategy, and then responding~\citep{LEPPER2002135}, many tutoring models perform all of them in one forward pass.
Recent studies~\citep{macina-etal-2023-opportunities, macina2023mathdial} have shown that this can lead to several deficiencies that can be detrimental to student learning, for example, in math tutoring. %
Despite impressive performance on math reasoning benchmarks~\citep{cobbe2021training,hendrycksmath2021}, dialog tutors often generate hallucinated outputs and present erroneous information to students, for example, because they assess an incorrect solution as correct.
We show an example of this in~\Cref{fig:overview}.

In this paper, we alleviate this problem by decoupling the verification of student solutions from response generation with a modular approach.
As opposed to the common approach, the model does not directly generate the tutor response from the students' utterances, whereby solution assessment is done implicitly in the models' activations, but rather receives the output of an additional verification model that assesses solutions 
and can therefore also be more specialized.
We hypothesize that this increases the correctness of the model as well as makes the response more targeted to the error because the response generation module is already aware of the exact student error.
Furthermore, this architecture more closely mimics human tutors.

To test our approach and train verifiers, we collect a dataset of ca. 1k student solutions and their stepwise reasoning chains in the domain of multi-step math problem-solving, which will be released publicly.
This dataset augments the math dialog tutoring corpus MathDial~\citep{macina2023mathdial}, which we use for evaluating dialog tutoring models, by teacher-annotated verifications of the first erroneous step in the student solution (\Cref{sec:data-collection}). 

We propose three verification approaches based on prompting and finetuning language models.
Besides a simple classification-based approach for verification, we also generate a textual verification and notably align student solution steps to steps of a reference solution (\cref{sec:verifiers}) to verify the student solution.
We find that using our data for finetuning helps smaller LLMs surpass prompted state-of-the-art LLMs.
Furthermore, incorporating the verification output in the response generation step (\cref{sec:response_generation_method}) clearly improves their performance in terms of both extensive automatic (\cref{sec:results}) and human evaluation using real teachers (\cref{sec:human_eval}): the generated responses are more targeted to the exact student error, there are less hallucinations, and there is more actionable scaffolding feedback for the student.
In general, we find that such improvements are much stronger when the verification output is correct (\cref{sec:ablation-alignment}) indicating a large potential for the community to improve dialog tutors by adding verifiers.

\section{Background \& Related Work}
Dialog tutoring aims at building models that can tutor human students through a conversation.
For example, dialog tutoring has been proposed for second-language acquisition~\citep{stasaski2020cima, caines2020teacher, kwon2024biped}, to answer questions in science~\citep{chevalier2024sciencetutors}, or to solve math word problems (MWPs)~\citep{macina2023mathdial}.
In each case, the model should guide the learner to solving a problem (e.g. the MWP or translation of a phrase) not by telling the solution outright, but rather by using scaffolding techniques that give students space for guided exploration and self-correction.
For example, the tutor might elicit the students' thinking by asking a question that challenges their understanding of the problem~\citep{reiser2004,anghileri2006scaffolding}.

Capturing such intricate tutoring strategies is hard and requires teachers years to master.
Due to this complexity, %
most previous dialog tutoring systems were human-authored, notably the AutoTutor family~\citep{nye2014autotutor}, LISP tutor~\citep{anderson1985intelligent} which uses a large set of rules to verify student programming solutions, or any systems built using CTAT which requires enumerating all possible solutions or writing complex production rules~\citep{aleven2016example}.
However, scaling such human-authored systems can quickly explode in both complexity and human effort~\citep{macina-etal-2023-opportunities}.
Due to this and rapid progress in language generation from learning large models based on large amounts of data, LLMs such as LearnLM~\citep{learnlm2024} have recently become popular in favor of 
human-authored, rule-based systems.

\paragraph{Problem formulation}
Formally, the goal of dialog tutoring is to continue a tutoring dialog consisting of a sequence of $T-1$ turns $\mathcal{H} \coloneqq (\mathbf{u}_1, \dots, \mathbf{u}_{T-1})$ taken by either student or teacher and where $\mathbf{u}_t \in \mathcal{V}^\ast$ is constructed from a fixed model vocabulary $\mathcal{V}$.
Continuation then means generating a new utterance $u_T \in \mathcal{V}^\ast$ that follows the above principles.
Often there is also background knowledge that is either required or helpful to tutor a certain concept, such as grammar rules~\citep{stasaski2020cima}, or textbook knowledge~\citep{wang2024book2dial}, and can be used to ground $\vu_T$.
In this work, we deal with teaching math word problem-solving and therefore use textual knowledge $\vk \in \mathcal{V}^\ast$, for instance, the math word problem 
and background knowledge like math rules.

\paragraph{Tutor models}
Using such data, a simple approach is generating the teacher response directly by pairing the following model, parameterized by weights $\vparam$, with a decoding algorithm, such as beam search or greedy decoding:
\begin{align}
\label{baseline_model}
    p_{\text{\vparam}}(\mathbf{y} \mid \mathcal{H}, \mathbf{k}) &= \prod_{i=1}^{|\mathbf{y}|} p_{\text{\vparam}}(\mathbf{y}_i \mid \mathbf{y}_{< i}, \mathcal{H}, \mathbf{k}) \text{.}
\end{align}
This model is straightforward to implement and learn from data but prior work has shown that it suffers from generating factually incorrect outputs~\citep{macina2023mathdial}.
Therefore, in this work we break down response generation into two steps: verification, where it is first assessed whether the student solution is correct, and generation.

\paragraph{Verification}
Verification is challenging in its own right and has recently been tackled for general reasoning problems.
For example, ROSCOE~\citep{golovneva2023roscoe} presents unsupervised metrics to assess the correctness of a models' chain-of-thought (CoT) reasoning, and~\citep{jacovi2024chain} evaluate open-domain question answering for logical errors.  
The outputs of verifiers have subsequently also been used for self-refinement of LLMs~\citep{madaan2023selfrefine, shinn2024reflexion} and also allow targeted feedback for the training of student LLMs with teacher LLMs~\citep{saha2023can, wang2024tpd}.
Closely related to our work~\citep{wang-etal-2024-bridging} define broad error categories, such as miscalculation, to understand the cause of incorrect reasoning by students and condition on it to generate teacher responses.
We compare to this baseline and call it {\bf Error Reason}.

\section{Verification-based Response Generation}
We first introduce the task of verification and different verifiers in~\cref{sec:verifiers}.
Afterwards, in~\cref{sec:response_generation_method}, we combine verification and response generation for modular tutor response generation.

\subsection{Verification}\label{sec:verifiers}
We deal with the verification of student solutions to a given math word problem $\vq \in \mathcal{V}^\ast$.
The solutions can be described by a sequence of substep solutions $\studentsol=\{\vs_1, ..., \vs_N\}$, where each $\vs_n \in \mathcal{V}^\ast$ and $\vs_N$ is the final solution.
Usually, $\studentsol$ is described by the student in one of the student utterances $\vu_t$.
The task of the model is to assess whether $\vs_N$ is the correct solution to $\vq$ and if not, potentially, to identify which step $\vs_n$ caused the error.
Oftentimes, this can be done by comparing to a reference solution $\correctsol=\{\widehat{\vs}_1, ..., \widehat{\vs}_M\}$ that is either given or model-generated and might differ in length. 
All verifiers which we discuss next can then be described by a learned function $\smash{v_{\text{\vparam}^\prime}(\vv \mid \studentsol, \correctsol)}$, usually an LLM.
Here, $\vv$ is the verification output and $\correctsol$ may be an empty string if no reference solution is given.
In the following, we introduce different verifiers.

\paragraph{\textcolor{mygreen}{Classification-based Verification}}
A comparably simple approach to verification is classifying whether the student solution ${\vs}_{\text{\vq}}$ is correct
using a binary classifier.
We call this {\bf Overall Verification}.
Similarly, identifying the first error step $\vs_{n}$ can be framed as multi-class classification with labels $\{0, \dots, N\}$, where $0$ means no mistake. We call this {\bf Stepwise Verification}.
Alternatively, Stepwise Verification (iterative) can be framed as a binary classification for each step  $\vs_n$ whether it is correct. The first error step is the first step classified as error.

\paragraph{\errordescription}
While conceptually easy, classification-based approaches locate the first error without explaining the exact issue.
Therefore, we propose to use an LLM to directly describe the error, and the concrete first error step, in a textual format, and call this {\bf Error Description}.
For this, we prompt the LLM with the prompt outlined in~\Cref{app:prompts}.
In comparison to~\citet{wang-etal-2024-bridging}, this error description is allowed to be free-form and does not map to predefined error types.
The LLM-generated error step description can then be passed to a tutor response generation model.

\begin{algorithm}[t]
    \footnotesize
  \caption{Modified Needleman-Wunsch.}
  \label{alg:needleman-wunsch}
  \begin{algorithmic}[1]
    \Require{Solution attempt $\vs_{\text{\vq}} = \{\vs_1, ..., \vs_N\}$}
    \Require{Reference solution $\widehat{\vs}_{\text{\vq}} = \{\widehat{\vs}_1, ..., \widehat{\vs}_M\}$}
    \Require{Gap cost $c$, similarity threshold $t$}
    \Ensure{Optimal alignment of $\vs_{\text{\vq}}$ and $\widehat{\vs}_{\text{\vq}}$}
    \State $F \gets \text{zeros\_matrix}(M+1, N+1)$ \Comment{initialize}
    \State $F[1:M+1, 0] \gets [i \cdot c \text{ for } i \text{ in } 1 \ldots M]$
    \State $F[0, 1:N+1] \gets [i \cdot c \text{ for } i \text{ in } 1 \ldots N]$
    \For{$i \gets 1 \text{ to } M$}   %
        \State $\ve_{\widehat{\vs}_m} \gets \text{embed}(\widehat{\vs}_m)$
        \For{$j \gets 1 \text{ to } N$}
        \State $\smash{\ve_{{\widehat{\vs}_n}} \gets \text{embed}({\vs_n})}$
        \State $\smash{F[i, j] \gets \text{cosine\_similarity}(\ve_{{\widehat{\vs}_m}}, \ve_{{{\vs_n}}})}$
            \If{$F[i, j] \geq t$} \Comment{exact match}
                \State $F[i, j] \gets F[i-1, j-1] + F[i, j]$
            \Else  \Comment{near match or gap}
                \State $F[i, j] \gets \max(F[i-1, j-1] - 1 + F[i, j], F[i-1, j] + c, F[i, j-1] + c)$
            \EndIf
        \EndFor
    \EndFor
    \State $\va = \{(\va_1, \widehat{\va}_1), ..., (\va_L, \widehat{\va}_L)\} \gets \text{backtrack}(F, \vs_{\text{\vq}}, \widehat{\vs}_{\text{\vq}})$
    \State \textbf{return} Globally-optimal alignment $\va$
    \end{algorithmic}
    \label{alg:nw}
\end{algorithm}
\paragraph{\textcolor{myorange}{(Step) Alignment}}

As our third verification approach,
we align the steps in the student's solution with a reference solution, and
compare the steps in the student and reference solution %
to localize errors. 
We call this approach {\bf Step Alignment}. %
As the order of steps in the solutions matters, a greedy algorithm that finds the most similar steps across the two solutions is insufficient. 
Thus, we frame verification as a sequence alignment problem.

The input to the alignment algorithm is the student solution $\studentsol$ with $N$ steps and the reference solution $\correctsol$ with $M$ steps.
Note that here we are aligning solution steps which can be long strings. This is different from other sequence alignment problems in NLP, where typically tokens are aligned~\citep[inter alia]{paolini2021structured}.
The output is a sequence of tuples $\{(\va_1, \widehat{\va}_1), ..., (\va_L, \widehat{\va}_L)\}$ of length $L$, where each $\va_l$ and $\widehat{\va}_l$ can be either a step of $\studentsol$ and $\correctsol$, respectively, or a special symbol $\oslash$. 
Here, $\va_l = \oslash$ indicates a missing step in the student solution ($-$) and $\widehat{\vs}_l = \oslash$ indicates an additional step ($+$). 

In our implementation, we use the Needleman-Wunsch (NW) algorithm~\citep{needleman1970general} as it guarantees an optimal alignment with respect to a chosen cost function. We use a modified version of the algorithm for semantic sequence alignment and use sentence embeddings~\citep{reimers-2019-sentence-bert} to measure the similarity between steps.
We detail our adaptation of the NW algorithm in~\cref{alg:nw} and describe each step in the following.
The NW algorithm iterates over all possible pairs of substeps from $\studentsol$ and $\correctsol$ and calculates a cost for each pair.
Since each substep is a string, we use semantic string similarity measured by the cosine similarity of the contextual embeddings of the substeps.
In our experiments, this performed better than just matching the final numerical solution of the substeps (cf.~\cref{sec:ablation-alignment} for results and a comparison of embedding models).
As not all high sentence embedding scores indicate a significant match, we introduce a threshold $t$ to differentiate between exact and near matches. If the similarity is higher than a threshold $t$, the pair is deemed as an exact match and its similarity is added to the similarity of their predecessors.
If it is smaller than $t$ it could still be a near match if the sequence similarity is high enough after incurring a penalty ($-1$). 
The last option is a gap if the sum is larger than adding a predefined gap cost $c$ to either a pair of the previous student and current reference solution step, or a pair of the current student and previous reference solution step.
Altogether, this forms a similarity matrix $F$ of size $(N+1) \times (M+1)$.
The alignment is finally found by backtracking (moving only to adjacent entries with each step) from entry $\smash{F_{N+1, M+1}}$.

Similar to the classification-based approach, the alignment output can not directly be used in a response generation model but has to be converted to a formatted verification output string.
For this, we use a preformatted template shown in~\cref{app:prompts}.
The template groups together the missing, additional and matching steps to produce $\vv$ from the alignment produced by the NW algorithm.

\subsection{Response Generation}
\label{sec:response_generation_method}
Direct generation of tutor responses can be challenging because one model has to reason over the student solution, pick a teaching strategy, and generate a response in one step.
This has been shown to produce hallucinations~\citep{macina2023mathdial}.
We tackle this by incorporating an additional verification step that informs the response generation model, as previously discussed.
Our aim is to split the task into two less complex tasks which should reduce errors if each task can be performed well enough and has been shown to reduce hallucinations in document-grounded dialog~\citep{adolphs-etal-2022-reason} and question answering~\citep{press-etal-2023-measuring}.

The verifier and response generation model are combined in a two-stage approach.
First, the verifier outputs a verification $\vv$ of the student solution $\smash{\studentsol}$ based on a reference solution $\smash{\correctsol}$. 
Then, the response generation model is conditioned on $\vv$, the dialog history $\mathcal{H}$, and background knowledge $\vk$.
In our work, $\vk$ consists of the student solution $\studentsol$, optionally the reference solution $\correctsol$, and the math word problem $\vq$.
If $v_{\text{\vparam}}$ is a distribution over verification labels, the overall model is: \begin{equation}
\begin{split}
    &p(\mathbf{y}, \vv \mid \mathcal{H}, \mathbf{k}) = \underbrace{v_{\text{\vparam}^\prime}(\vv\mid\studentsol, \correctsol)}_{\text{verification}}\cdot\underbrace{p_{\text{\vparam}}(\mathbf{y} \mid\mathcal{H}, \mathbf{k}, \vv)}_{\text{generation}} 
\end{split}
\end{equation}
The full model provides us with a verification output and the generated response which makes the internal reasoning of the tutor model in terms of student errors more explicit and controllable.

\section{Data Collection}\label{sec:data-collection}
We propose and evaluate various verifiers in this work.
Since some of them require training data and to evaluate their performance, we collect a dataset of 1,002 human-produced verification outputs to train and evaluate them.
This is similar in size to a related corpora~\citep{jacovi2024chain}.
In this section, we describe the annotation task and data collection.

\paragraph{Incorrect Student Solutions Source}
Our work extends MathDial~\citep{macina2023mathdial} by having teachers annotate incorrect student solutions from the dataset with their first error step. There, these incorrect student solutions were used to condition a student model (\textsc{InstructGPT}) to generate responses in a dialogue with a human teacher.

Specifically, these problems are based on the GSM8k~\citep{cobbe2021training} dataset of multi-step math word problems. In MathDial, the reasoning chains are generated using a 2-shot CoT prompt with \textsc{gpt-3.5-turbo}, and temperature sampling ($T=0.7$) is used to
get multiple reasoning paths ($n=50$). Finally, the most common incorrect solution is chosen. Subsequently, their student model is prompted to respond to a human teacher as a student who tries to solve a problem with a particular incorrect solution. 

To not skew our dataset to errors, we balance it with rephrased reference solutions from the student model. We reproduce the student model prompt from MathDial to generate student responses using the reference solutions. All reference solutions and student responses with incorrect solutions are part of the dataset. Details are in~\Cref{appendix:step-errors-collection}.

\paragraph{Student Solution Annotation}
The objective of the annotation is to mark the exact step of the first error in the student solution. 
We do not annotate error steps after the first one to decrease ambiguity, as they frequently stem from the first error. 
We recruit teachers through Prolific,
who first read the problem and then mark the precise step of the first error in the student solution. 
Teachers can access the reference solution to reduce task complexity. Details of the task, the user interface, and examples of collected data are in~\Cref{appendix:step-errors-collection}.
To compute agreement, 10\% of the samples are annotated by one additional annotator with an 
inter-rater reliability of Cohen's $\kappa=0.75$ indicating substantial agreement~\citep{cohen1960coefficient}. %
We show the distribution of incorrect student solution steps in~\Cref{fig:dataset-error-step}.

\section{Experiments}
We evaluate different verifiers on our dataset and use them to inform response generation models to improve their correctness.
Since we extend MathDial with additional annotations we use MathDial dialogues for evaluating tutor response generation.
Besides math problem and student solution in a dialog, we either use a model-generated CoT reference solution if marked by ``solution'' or no reference solution as input to the models.
Next, we detail metrics and models.

\subsection{Metrics}
For teacher response generation, we evaluate the generated output $\vu_T$ of each model by comparing it to a human-annotated response $\smash{\widehat{\vu}_{T}}$ from MathDial.  We report standard text generation metrics: the sacrebleu~\citep{post-2018-call} implementation of BLEU (sBLEU) to measure word overlap and BERTScore~\citep{Zhang2020BERTScore} (BF1, using the \textit{all-MiniLM-L6-v2} checkpoint) to measure semantic similarity. Moreover, we report the knowledge F1 (KF1) score with respect to the grounding information (correct solution in the case of MathDial) which has been used as a proxy for faithfulness in prior work~\citep{daheim2023elastic}. 
Similar to~\citep{zheng2024judging, learnlm2024}, we also prompt LLAMA3-70B and use it complementary to human evaluation (the same task and instructions are used in both) to assess
how \emph{targeted}, \emph{correct}, and \emph{actionable} a response is.
Details about the LLM-based evaluation are found in~\cref{app:llm-eval}.

\subsection{Models}
For both verification and response generation, we use different prompted or finetuned models.
For verification, we compare the closed-source model GPT-3.5 to the open models LLAMA2 and LLAMA3. For the latter, we prompt the 70B version of the models and finetune LLAMA2-7B using LoRA.
For response generation, we evaluate prompting GPT-3.5 and finetuning the encoder-decoder model Flan-T5 with 3B parameters.
We finetune Flan-T5 again using LoRA for both the direct modeling and verify-then-generate approach.

\begin{table}
    \centering
    \resizebox{\linewidth}{!}{\begin{tabular}{lccc|c}
        Model & \multicolumn{3}{c}{{\color{mygreen}Overall Verification}} & {\color{mygreen}Stepwise}\\
         & Corr. F1 & Err. F1 & F1 & micro F1 \\
        \hline
        \multicolumn{5}{l}{\rule{0pt}{0.1in}\bf Few-shot} \\
        \;GPT3.5 & 0.66 & 0.52 & 0.59 & 0.42\\
        \;\; + solution & {\bf 0.97} & {\bf 0.97} & \textbf{0.97} & 0.61 \\
        \;Llama2-70B & 0.69  & 0.38  & 0.54 & 0.17 \\
        \;\; + solution & 0.78  & 0.59  & 0.68 & 0.48\\
        \;Llama3-70B & 0.74  & 0.58  & 0.66 & \textbf{0.56}\\
        \;\; + solution & 0.90  & 0.87  & 0.89 & \textbf{0.70} \\
        \multicolumn{5}{l}{\rule{0pt}{0.1in}\bf Finetuned} \\
        \;Llama2-7B & {\bf 0.89} & 0.67 & 0.78 & 0.20\\
        \;\; + solution & 0.81 & {\bf 0.80} & {\bf 0.80} & 0.28\\
        \hline
    \end{tabular}}
    \caption{Verifying student solutions can be challenging even for strong LLMs.  Models are worse at verifying erroneous responses (Err. F1) than correct responses (Corr. F1). Providing a reference solution improves all models. Fine-tuning using our data can make models more robust when no such solution is present and make small models outperform larger prompted ones.}
    \label{tab:error-finding-math-table}
\end{table}
\begin{table*}[t!]
    \centering
    \resizebox{.9\linewidth}{!}{\begin{tabular}{llcccccc}
        &  & \multicolumn{3}{c}{Automatic Metrics} & \multicolumn{3}{c}{LLM Judge (\%)} \\
        Model & Variant & sBLEU & KF1 & BF1 & Targeted & Correct & Actionable  \\
        \hline
        - & Human & 100.0 & 100.0 & 100.0 & 27 & 82 & 87 \\
        \hline
        \multicolumn{8}{l}{\rule{0pt}{0.1in}\bf Few-shot}\\
        \, GPT-3.5 & Baseline & 2.0 & 27.0 & 51.2 & 29 & 37 & 27 \\
         & {\bf Error Reason} & 1.5  & 22.5  & 46.7  & 34 & 40 & {\bf 56} \\
         & {\errordescription} & 2.8 & \textbf{30.3}  & 52.6  & \textbf{62} & {\bf 66} & 45 \\
         & {\color{myorange} Step Alignment} & 2.3 & 29.8  & {\bf 53.3}  & 42 & 61 & 26 \\
        \multicolumn{8}{l}{\rule{0pt}{0.1in}\bf Finetuned} \\
        \, Flan-T5-3B & Baseline & 2.6 & {\bf 27.6} & \textbf{56.0} & 1 & 89 & 76 \\
         & {\errordescription} & \textbf{3.0} & 26.7 & \textbf{56.0} & {\bf 2} & \textbf{92} & \textbf{84} \\
        \hline
        
    \end{tabular}}
    \caption{
    Adding an additional verification stage to ground tutor response generation models leads to responses that are more targeted at the student error, less frequently hallucinated, and more actionable for the student, both for finetuned and prompted models.
    Proving a textual {\errordescription} of the student solution performs better than {\color{myorange} Step Alignment} of student and reference solution, as well as providing a shorter {\bf Error Reason}.
    }
    \label{tab:response_generation}
\end{table*}

\section{Results}\label{sec:results}
We first show the performance of different verification models in~\Cref{sec:error_finding} and then use verification models in response generation in~\Cref{sec:response_generation}.

\subsection{Verification}
\label{sec:error_finding}
In this section, we benchmark LLMs on their ability to evaluate the correctness of student solutions using the {\color{mygreen} Overall Verification} and {\color{mygreen} Stepwise Verification} approaches from~\cref{sec:verifiers}.
For {\color{mygreen} Stepwise Verification} we use the multi-class classification approach, because it performed better than iterative classification in our experiments. 
A comparison is found in~\cref{appendix:verification-details}.
We measure the F1 score (balanced dataset), in particular, micro F1 for {\color{mygreen} Stepwise Verification} (imbalanced dataset, see~\Cref{fig:dataset-error-step}).
We find in~\Cref{tab:error-finding-math-table} that {\color{mygreen} Overall Verification} can be challenging even for state-of-the-art LLMs.
All prompted models show comparably low performance when prompted without a reference solution and especially struggle with identifying incorrect responses.
Providing a reference solution improves results significantly. 
However, for {\color{mygreen} Stepwise Verification} even the reference solution does not improve micro F1 beyond 0.70. This result is consistent with expert educator-based assessment~\citep{yen-hsu-2023-three} and LLM self-correction results~\citep{huang2024large}. %

Interestingly, our dataset can be used effectively for finetuning.
Even a smaller LLAMA2-7B model can outperform larger prompted models on {\color{mygreen} Overall Verification}, especially when no solution is provided.
Potentially, the additional finetuning steps make it easier for the model to also solve the problem before verification.
The finetuned {\color{mygreen} Stepwise Verification} model outperforms its larger prompted counterpart LLAMA2-70B when no solution is provided.
Results for finetuning show a ten-fold cross-validation.
Further details are in~\Cref{finetuning-appendix}.

\subsection{Response Generation}
\label{sec:response_generation}
Next, we show in~\Cref{tab:response_generation} that combining verification and tutor response generation models can improve the quality of the generated responses.
We compare the {\errordescription} and {\color{myorange} Step Alignment} verifiers to direct response generation and using the {\bf Error Reason}~\citep{wang-etal-2024-bridging}.
There, the error is categorized into either: \emph{guess}, \emph{misinterpret}, \emph{right-idea}, \emph{imprecise}, \emph{not sure}, or \emph{careless}.
We use a subset of MathDial, where the student describes their solution to the teacher in the dialog, and generate the following teacher utterance.

First, we prompt GPT-3.5 using the prompt templates from~\Cref{sec:verifiers} for comparability.
We find that providing only the {\bf Error Reason} does not improve over the direct baseline in simpler automatic metrics (sBLEU, KF1, BF1) but only in terms of the LLM-based judging.
Using the more detailed {\errordescription} which provides the exact mistake of the student gives larger improvements, both in terms of automatic metrics and LLM-based judging.
Similarly, we find  {\color{myorange} Step Alignment} to be helpful, but to provide less actionable responses.
When finetuning with the {\errordescription}, we obtain improvements over the finetuned baseline but they are smaller and do not hold for each metric. %

Our qualitative analysis shows that both {\color{myorange} Step Alignment} and {\errordescription} result in responses that better localize the exact student error.
For example, the baseline often assesses the solution wrongly or skips the first error step and instead asks for the solution of a later step.
Examples are shown in~\cref{tab:gpt-generated-examples} and \cref{tab:finetuned-ex-1}.
\Cref{sec:human_eval} confirms our results by human evaluation.

\begin{table}[t]
    \centering
    \resizebox{\linewidth}{!}{\begin{tabular}{llccc}
        Model & Variant & target$\uparrow$ & corr$\uparrow$ & act$\uparrow$ \\
        \hline
        \multicolumn{5}{l}{\rule{0pt}{0.1in}\bf Verification} \\
        \;GPT3.5 & {\color{myblue} Error Description} & - & 70.6 & -  \\
        \;Llama3-70B & {\color{myblue} Error Description} & - & 82.4 & -  \\
        \hline
        \multicolumn{5}{l}{\rule{0pt}{0.1in}\bf Response Generation} \\
        - & Human & 35.0 & 45.0 & 42.5  \\
        \multicolumn{5}{l}{\rule{0pt}{0.1in}\bf Few-shot} \\
        \;GPT3.5 & Baseline & 30.0 & 37.5 & 30.0  \\
        & {\bf Error Reason} & 27.5 & 22.5 & 37.5  \\
        & {\color{myblue} Error Description} & \textbf{57.5} & \textbf{62.5} & \textbf{45.0}  \\
        & {\color{myorange} Step Alignment} & \textbf{57.5} & \textbf{60.0} & 27.5  \\
        \multicolumn{5}{l}{\rule{0pt}{0.1in}\bf Finetuned} \\
        \;Flan-T5-3B & Baseline & 7.5 & 20.0 & 25.0  \\
        & {\color{myblue} Error Description} & {\bf 20.0} & {\bf 35.0} & {\bf 35.0}  \\
        \hline
    \end{tabular}}
    \caption{Human evaluation with four expert annotators shows that verification before generation improves along the targetedness, correctness, and actionability (without telling the solution) of responses.
    We find that {\errordescription} works best and improves both prompted and finetuned models.
    \label{tab:human-evaluation-quality}}
\vspace{-5mm}
\end{table}

\subsection{Human Evaluation}
\label{sec:human_eval}
We conduct a human evaluation using teachers as expert annotators.
All annotators are recruited on Prolific after manual screening.
We assess whether the generated responses are \emph{targeted}, \emph{correct}, and \emph{actionable} without outright telling the solution. 
Annotators are blind to the model source. The exact questions are as follows. 
1) (\textit{Targeted (T)}) Does the Teacher point out the root cause of the student's mistake? 
2) (\textit{Correctness (C)}) Is the Teacher's response factually correct with respect to the reference solution?
3) (\textit{Actionable (A)}) - Does the Teacher provide actionable steps to let the Student correct the mistake without giving away the full answer?
More details are in~\Cref{appendix-human-eval}.

Responses from 6 models and one human response from MathDial were annotated for a random set of 40 conversations.
To compute inter-rater reliability, 9 conversations were annotated with at least 2 raters for each response. Cohen's kappa is 0.21, 0.25, and 0.13 for targeted, correctness, and actionable. For {\errordescription} correctness it is $\kappa=0.30$. 
Next, we describe the results, first on verification and then for response generation.

\paragraph{Verification}

Annotators assess the {\errordescription} as correct if the exact mistake of the student is found and incorrect when the model says that the solution is correct when it is not and vice versa, misses the step of the error, or is generic.
Results in~\Cref{tab:human-evaluation-quality} show that LLAMA3-70B outperforms GPT-3.5 but also with $82.4\%$ of the errors being found correctly.

\paragraph{Response Generation}
Next, in~\Cref{tab:human-evaluation-quality}, we evaluate how {\emph{targeted}}, \emph{correct}, and \emph{actionable} the responses generated by different models are.
We find that providing the {\bf Error Reason} improves over the baseline only in terms of how actionable responses are.
We hypothesize that conditioning on only the reason is insufficient for a targeted response.
{\errordescription} and {\color{myorange} Step Alignment} provide more information regarding the exact error and therefore improve strongly over the baseline in both targetedness and correctness. 
Using {\color{myorange} Step Alignment} information also does not improve actionability but {\errordescription} improves it.
The same improvements also hold for using the {\errordescription} for a finetuned model.
All in all, we find strong evidence that using our verify-then-generate approach improves teacher response generation.

\section{Ablation Studies}
Next, we provide further ablations, first on the cost function used in the NW algorithm (\cref{sec:ablation-alignment}) and then on the impact of verification before response generation based on its correctness and problem difficulty (\cref{sec:ablation-verification}).

\subsection{Alignment}
\label{sec:ablation-alignment}
\begin{table}[t]
    \centering
    \centering
    \begin{tabular}{lccc}
         {\color{myorange} Step Alignment} & $t^\ast$ & $c^\ast$ & Accuracy(\%) \\
        \hline
        Random & 0.6 & -0.2 & 43.6 \\
        Solution Match & 0.6 & -0.3 & 51.9 \\ 
        \hline
        SBERT & 0.8 & -0.1 & 58.2 \\
        Roscoe & 0.9 & -0.1 & 61.4 \\
        \hline
    \end{tabular}
    \caption{
    We compare different cost functions for {\color{myorange} Step Alignment} with the Needleman-Wunsch algorithm based on 30 human-annotated alignments.
    Semantic-similarity-based cost function (SBERT, Roscoe) performs better than random cost or an indicator function of whether the numerical substep solutions match. 
    \label{tab:alignment-eval}
    }
\end{table}
We compare different cost functions used for the NW {\color{myorange} Step Alignment} algorithm in~\cref{tab:alignment-eval}.
For the comparison, 30 alignments between a student and reference solution were produced by humans and the accuracy of student solution step alignment is measured.
As cost functions we use the cosine similarity of Sentence-BERT (SBERT)~\citep{reimers-2019-sentence-bert} embeddings and embeddings from a model trained on Roscoe~\citep{golovneva2023roscoe}, as well as a random cost and an indicator function that is $1$ when two substeps have the same numerical solution and $0$ otherwise.
Similarity threshold $t$ and gap cost $c$ are optimized via a hyperparameter grid search, as indicated by $t^\ast$ and $c^\ast$.
We find that cosine similarity works best and training on relevant math data fine-tuning in Roscoe further improves performance.

\subsection{Verification}
\label{sec:ablation-verification}
\paragraph{Verification correctness is important} 
\begin{table}
    \centering
    \begin{tabular}{llccc}
        Model & {\color{myblue} Error Desc.} & T & C & A \\
        \hline
        \multicolumn{5}{l}{\rule{0pt}{0.1in}\bf Few-shot} \\
        GPT3.5 & incorrect & 50.0 & {\bf 62.5} & {\bf 50.0}  \\
         & correct & {\bf 62.5} & {\bf 62.5} & 41.7  \\
        Llama3-70B & incorrect & 41.7 & 33.3 & {\bf 25.0}  \\
         & correct & {\bf 82.1} & {\bf 71.4} & 21.4  \\
        \multicolumn{5}{l}{\rule{0pt}{0.1in}\bf Finetuned} \\
        Flan-T5-3B & incorrect & 08.3 & 25.0 & {\bf 41.7}  \\
         & correct & {\bf 25.0} & {\bf 39.3} & 32.1  \\
         \hline
    \end{tabular}
    \caption{We find that tutor responses are much more often correct and targeted if the {\errordescription} is correct. Data from human evaluation.
    \label{tab:correctness-ablation}}
    \vspace{-3mm}
\end{table}

We find in~\Cref{tab:correctness-ablation} that correct verification is important for subsequent response generation.
If it is correct, both targetedness and correctness are strongly improved over when it is incorrect.
However, actionability appears to be decreased which indicates less scaffolding and more teacher "telling the solution".

\paragraph{Problem difficulty influences verification} 
\begin{table}
    \centering
    \resizebox{\linewidth}{!}{\begin{tabular}{llcccc}
        Model & Steps & T & C & A & {\color{myblue} Error Desc.}\\
        \hline
        \multicolumn{6}{l}{\rule{0pt}{0.1in}\bf Few-shot} \\
        GPT3.5 & 3 & {\bf 62.5} & {\bf 75.0} &  37.5 & 50.0 \\
         & 4 & 61.1 & 55.6 & {\bf 55.6} & {\bf 72.2}\\
         & 5 & 50.0 & 64.3 & 50.0 & 50.0\\
        Llama3-70B & 3 & {\bf 87.5} & {\bf 75.0} &  12.5 & {\bf 87.5}\\
         & 4 & 72.2 & 72.2 & {\bf 72.2} & 72.2\\
         & 5 & 57.1 & 35.7 & 57.1 & 50.0 \\
        \multicolumn{6}{l}{\rule{0pt}{0.1in}\bf Finetuned} \\
        Flan-T5-3B & 3 & 12.5 & 25.0  & 25.0 & {\bf 87.5} \\
         & 4 & {\bf 27.8} & {\bf 50.0} & {\bf 50.0} & 72.2 \\
         & 5 & 14.3 & 21.4 & 14.3 & 50.0\\
         \hline
    \end{tabular}}
    \caption{
    For prompted models, responses for problems with shorter solution lengths are more often correct and targeted, because such problems are likely less complex. For finetuned models we do not find this trend. More steps can decrease description correctness ({\color{myblue} Error Desc.}). Data from human evaluation of {\errordescription}.
    \label{tab:difficulty-ablation}}
\end{table}

Finally, we show in~\Cref{tab:difficulty-ablation} that the performance of both verification and our verify-then-generate approach is heavily correlated with the number of reasoning steps that are used in the reference solution of a given math word problem.
We use this as a proxy for problem difficulty.
First of all, the performance of the LLAMA3 70B {\errordescription} decreases with the number of steps.
This is reflected in the decreased correctness and targetedness of the responses of the few-shot prompted LLAMA3 model.
For GPT-3.5 we do not find a similar conclusion for the {\errordescription} model but at least targetedness still decreases with the number of steps.
For the finetuned model we do not see similar trends but instead find the best performance for problems with four steps, likely because these are more common in the training data.

\section{Discussion \& Conclusion}
Student errors are key learning opportunities.
Tutors should recognize them and precisely guide students with targeted feedback without telling full solution. 
Motivated by effective teaching practice, we split the task of tutor response generation into two separate steps of verifying the student solution and generating a response.

To evaluate our approach, we collect a dataset of around 1k teacher-annotated solutions to augment an existing math tutoring corpus.
Our results show that splitting response generation into two steps can result in more targeted and correct responses that better scaffold human learning. %
We showcase this using both automatic evaluation and human evaluation annotated by teachers, both for prompted and finetuned models.

\section{Limitations}
\paragraph{Focus on scaffolding problem-solving} The tutoring scenarios which are considered are centered around the student problem-solving stage. In this case, students have prior knowledge. mostly understand the learning topics and practice them. However, different learning scenarios such as direct instruction, building rapport with students, or open-ended discussions are not considered in this work.

Evaluating student solutions and responding appropriately to a student's mistakes is inherently challenging, even for human teachers. 
Furthermore, teachers should ideally give adaptive feedback depending on the problem-solving strategy chosen by the student and treat different errors in different ways to uncover any misconceptions~\citep{nye2014autotutor}.
For example, in math, productive errors present important learning opportunities for students to learn from them~\citep{kapur2016examining,shaughnessy2021think,sinha2021problem}, e.g. by teacher-guided self-correction or targeted instruction, while unproductive errors, such as numerical miscalculations, could be easily resolved using a calculator~\citep{LEPPER2002135}.

\paragraph{Difficulty of obtaining student reasoning chains} Model-generated reasoning chains might contain the same biases as human students~\citep{opedal2024do}. On the other hand, there might be many additional differences from human student reasoning, e.g. students might not always stick to exact math notations or skip some steps in the explanations.  However, because such data from students is sensitive, we work with model-generated reasoning solutions and responses.

\paragraph{Focus on multi-step problems} Procedural or multi-step problems are the basis of most of the scientific disciplines, therefore we believe our approach should be general enough to work across any science subject, especially by including retrieval-augmented generation (RAG) from textbooks. However, it is still an open research question whether a similar solution would work for language learning or fact-based problems, and how models perform in languages other than English.

\paragraph{Evaluation is teacher-centered and complemented with an LLM-judge} Future work should focus on student user studies with AI tutors. However, this requires careful experimental consideration and safety mechanisms. Moreover, assessment of the responses is done exclusively by teachers and therefore future work should consider running assessments of the responses by students.

\section{Acknowledgements} 
This research work has been funded by a Swiss National Science Foundation award (\#201009), a Responsible AI grant by the Haslerstiftung, by the German Federal Ministry of Education and Research and the Hessian Ministry of Higher Education, Research, Science and the Arts within their joint support of the National Research Center for Applied Cybersecurity ATHENE. 
Nico Daheim acknowledges travel support from ELISE (GA no 951847).
Jakub Macina acknowledges funding from the ETH AI Center Doctoral Fellowship, Asuera Stiftung, and the ETH Zurich Foundation. 
We thank Sankalan Pal Chowdhury, Kumar Shridhar, Shehzaad Dhuliawala, and Justus-Jonas Erker for valuable feedback and discussions.

\section{Ethics Statement}
\paragraph{Intended usage} The benefits of our dataset are in understanding and designing AI technology to assist teachers and students during the problem-solving stage. 
Most importantly, the goal of such systems is to not replace human teachers, but rather enhance their capabilities and make them focus on important and human aspects of teaching. 
We will release the dataset under CC-BY-4.0 license~\footnote{\url{https://creativecommons.org/licenses/by/4.0/deed.en}} for further usage and exploration by the community.
This also adheres to the licensing of MathDial, which we extend.

\paragraph{Data Anonymization and Privacy} As the data in education are strictly confidential we obtained approval on the proposal\footnote{The study was approved by the ETH Zurich Ethics Committee (IRB) under EK-2024-N-97.} of the collection interface, questions and how long the data will be stored. All participants fill informed consent at the beginning of the annotation and may withdraw without reason at any time. We store only the necessary data and do not store any personally identifiable information. The collected data are stored anonymously and securely. Moreover, no student data are used in this work.

\paragraph{Accessibility and Potential Misuse} Our work focus on addressing hallucinations of LLM Tutors and their generation of factually incorrect responses. This directly addresses one of the important aspects of responsible use of AI which does not spread incorrect information, especially in education. We encourage the community to work on this important topic by open-sourcing the dataset, the code for running the benchmarks, and the methods used in this work. These are primarily intended for research purposes. As with any AI technology, the methods and dataset could be misused. However, we believe by open-sourcing the work we inform about the risks and capabilities of the technology a wider research community which then leads to further improvements.

\bibliography{anthology,custom}

\appendix

\section{Data Collection Details}\label{appendix:step-errors-collection}
The annotators are screened through Prolific to be teachers native in English with an overall acceptance rate of more than 98\% and with at least 500 submissions. We paid a minimum of \$20 per hour. Annotators are from the US, Canada, and the UK, with a balanced gender ratio, and their age range is from 25 to 53 years. All annotators have K12 experience and on average they have 12 years of experience in teaching. 

The annotators are first trained for the task with an interactive practice problem and then annotate student solutions. In one session one annotator performs 5 stepwise error verifications where they first pick the exact step with the error and then classify the error into 8 categories, with separate descriptions of the error for each category. We filter out all error descriptions not following the prescribed format to remove low-quality annotations.

\begin{figure}
    \begin{center}
        \resizebox{1.0\linewidth}{!}{\includegraphics{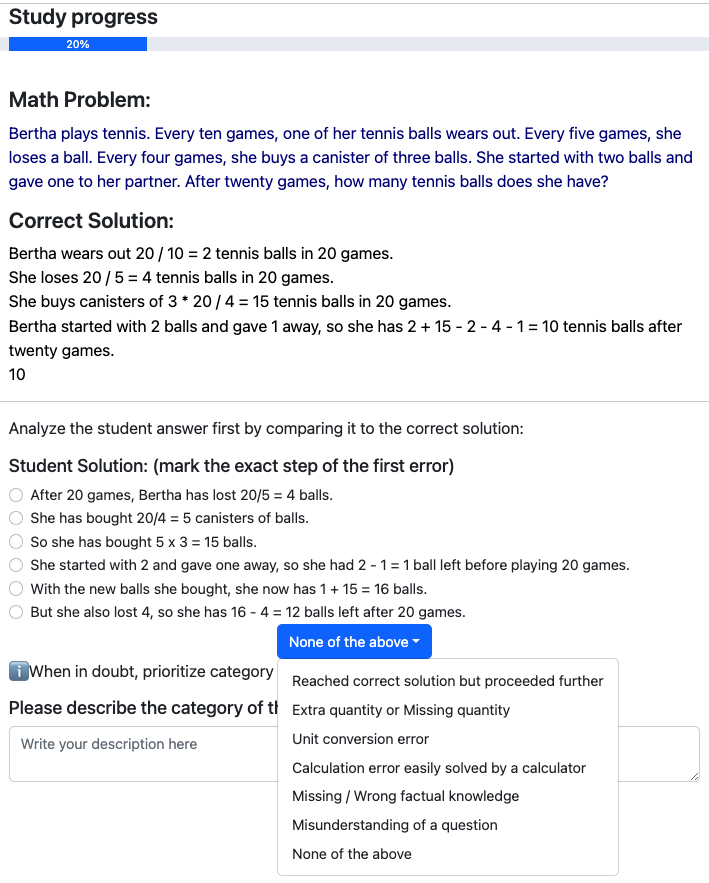}}
    \end{center}
    \caption{User interface for annotating the step of the first error, their categorization, and description of the error.
    \label{fig:interface-collection-ui}}
\end{figure}

The interface is shown in~\Cref{fig:interface-collection-ui}. The categories are: \textit{missing or incorrect factual knowledge, misunderstanding of the question, the reference solution reached but proceed further, missing quantity, extra quantity, unit conversion error, numerical calculation, other}. 

\subsection{Dataset Details}
\begin{figure}
    \begin{center}
        \resizebox{.95\linewidth}{!}{\includegraphics{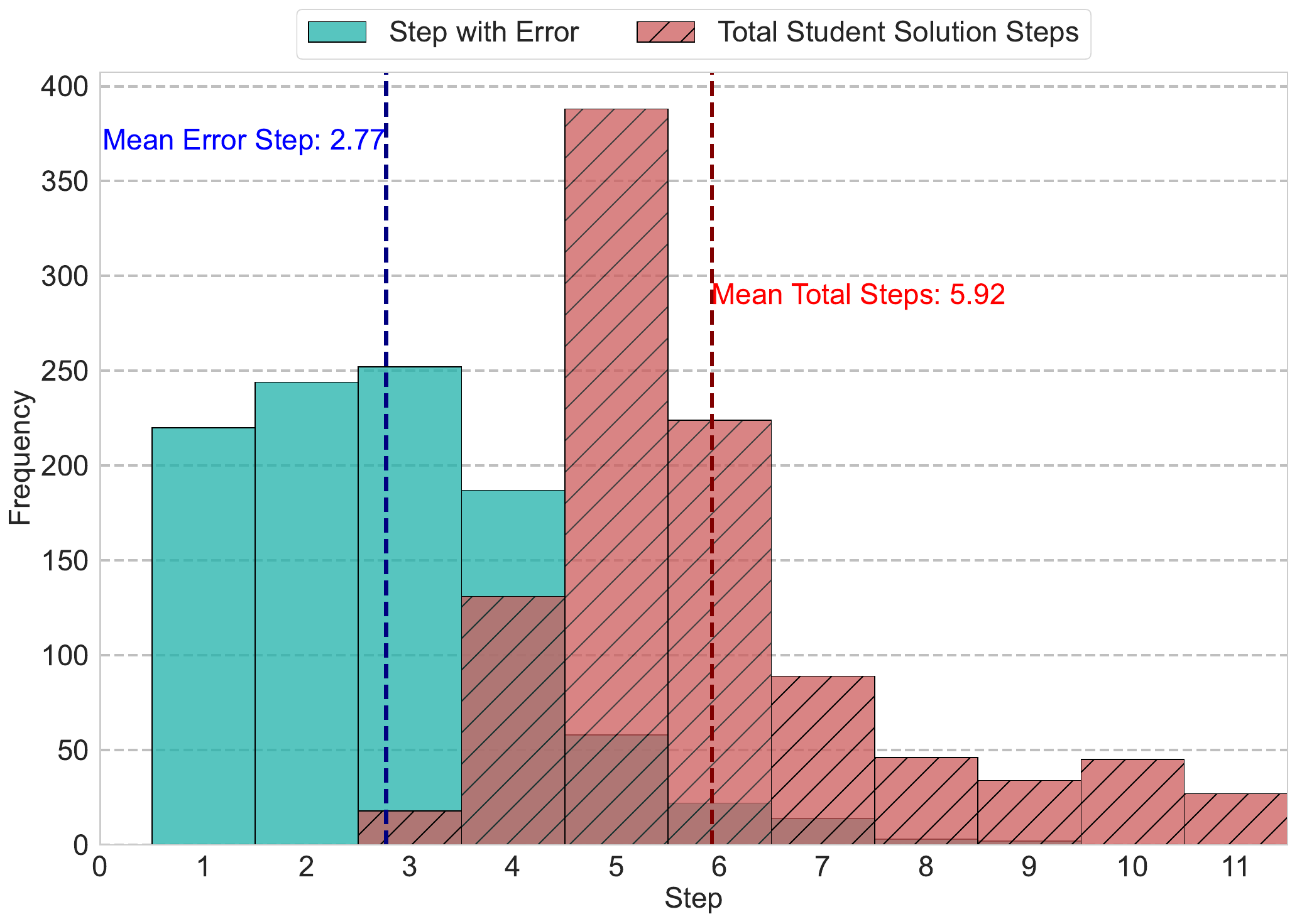}}
    \end{center}
    \caption{Dataset Distribution. The index of the step with the first error annotated by teachers and the total student solution steps.
    \label{fig:dataset-error-step} }
\end{figure}

The collected dataset is in English and from the domain of K12 math word problem-solving. Examples from the dataset are shown in~\Cref{{tab:dataset-examples}}. The dataset consists of 1002 data points with 612 unique math problems. The distribution of total student steps and the location of the first incorrect steps are shown in~\Cref{fig:dataset-error-step}. Notice the student solutions contain up to 11 steps with a mean of ca. 6 steps. The location of the first error ranges from 1 to 8 steps with majority of the errors between the first and third steps.

The incorrect student solutions and reference solutions are part of the MathDial dataset~\citep{macina2023mathdial}. The prompt used to generate correct student responses to balance the dataset with correct student responses are based on the Student model from~\citep{macina2023mathdial}.

\begin{table*}
\begin{tabularx}{\textwidth}{|X|}
\hline
\textbf{Problem:} Carl has been selling watermelons on the side of the road for \$3 each. This evening he went home with \$105 in profit and 18 watermelons. How many watermelons did he start out with this morning? \\
\textbf{Reference solution:} Carl sells his watermelons for \$3 each so today he sold \$105 / \$3 per watermelon = 35 watermelons. He had 18 watermelons left over, so this morning he started with 18 + 35 = 53 watermelons.\\
\textbf{Incorrect solution:} \textcolor{red}{Let's start by finding out how much money Carl made selling the 18 watermelons.} Since he sold each watermelon for \$3, he made 18 x \$3 = \$54. We know that his total profit for the day was \$105, so he must have started with \$105 - \$54 = \$51 worth of watermelons. Since each watermelon costs \$3, he must have started with 51 / \$3 = 17 watermelons. \\
\textbf{Student response:} \textcolor{red}{I started by finding out how much money Carl made selling the 18 watermelons.} Since he sold each watermelon for \$3, he made 18 x \$3 = \$54. Then I realized that his total profit for the day was \$105, so he must have started with \$105 - \$54 = \$51 worth of watermelons. Since each watermelon costs \$3, I concluded that he must have started with 51 / \$3 = 17 watermelons. \\
\textbf{Error category:} Misunderstanding of a question \\
\textbf{Error description:} Carl did not sell 18 watermelons, but 18 watermelons are left unsold. \\
\hline
\textbf{Problem:} Dakota gets hit by a bus and has to spend 3 days in the hospital. The hospital charges her \$900/day for her bed, \$250/hour for two specialists to look at her 15 minutes each, and \$1800 for the ambulance ride. How much is Dakota's medical bill? \\
\textbf{Reference solution:} First find the total cost of the hospital bed by multiplying the cost per day by the number of days: \$900/day * 3 days = \$2700. Then figure out how many minutes total the two specialists looked at Dakota: 15 minutes/specialist * 2 specialists = 30 minutes Then divide the number of minutes by the number of minutes per hour and multiply it by the specialists' hourly rate to find their total charge: 30 minutes / 60 minutes/hour * \$250/hour = \$125. Finally, add up the cost of the specialists, ambulance and bed to find Dakota's total bill: \$2700 + \$1800 + \$125 = \$4625\\
\textbf{Incorrect solution:} The hospital charges Dakota \$900 x 3 = \$2700 for her bed. \textcolor{red}{Each specialist charged her \$250/hour x 2 = \$500 for their 15 minutes each.} So, Dakota was charged \$500 x 2 = \$1000 for the two specialists. Therefore, her medical bill is \$2700 + \$1000 + \$1800 = \$5500 \\
\textbf{Student response:} I started by calculating the cost of the bed, which was \$900 x 3 days = \$2700. \textcolor{red}{Then I calculated the cost of the two specialists, which was \$250/hour x 2 specialists x 15 minutes each = \$500.} Then I added all the costs together to get the total cost of \$2700 + \$1000 + \$1800 = \$5500 \\
\textbf{Error category:} Misunderstanding of a question \\
\textbf{Error description:} Student computes charges for a full hour of 2 specialists, not just 15 minutes as indicated in the question. \\
\hline
\end{tabularx}
\caption{Examples from the collected dataset. The annotated error lines are in red. }
\label{tab:dataset-examples}
\end{table*}

\section{Details of \textcolor{mygreen}{Overall Verification} and \textcolor{mygreen}{Stepwise Verification}}\label{appendix:verification-details}
For \textcolor{mygreen}{Stepwise Verification}, we compare multi-class classification and iterative approach on our dataset and the results are in~\Cref{{tab:error-finding-iterative}}. The iterative approach classifies each step $\vs_{n}$ whether it is correct and therefore is more resource-intensive than multi-class classification. The multi-class classification directly predicts the label $\{0, \dots, N\}$ where 0 represents the solution is correct. The results indicate no improvements (with the exception of Llama2-70B) by using the iterative approach and in the main paper we therefore report multi-class classification results.

Moreover, to confirm the quality of our collected dataset, we run the same models on the smaller and simpler Roscoe human evaluation set~\citep{golovneva2023roscoe}. The dataset is smaller and contains 105 correct and 95 incorrect solutions. The results are shown in Table \ref{tab:error-finding-roscoe-dataset} and the conclusions are identical to our dataset.

\begin{table}
    \centering
    \begin{tabular}{lc}
        Model & micro F1\\
        \hline
        \;GPT3.5 multi-class & 0.42 \\
        \;\; + solution &  0.61 \\
        \;GPT3.5 iterative & 0.39 \\
        \;\; + solution &  0.36 \\
        \hline
        \;Llama2-70B multi-class &  0.17 \\
        \;\; + solution &  0.48\\
        \;Llama2-70B iterative &  0.36 \\
        \;\; + solution &  0.58\\
        \hline
        \;Llama3-70B multi-class &  \textbf{0.56} \\
        \;\; + solution &  \textbf{0.70}\\
        \;Llama3-70B iterative &  \textbf{0.56} \\
        \;\; + solution &  0.58\\
        \hline
    \end{tabular}
    \caption{Results of two approaches for \textcolor{mygreen}{Stepwise verification} and their micro F1 score. Multi-class classification directly predicts incorrect step $N$. On the other hand, the iterative approach iterates over each step and runs a binary prediction of whether the step is correct until the first incorrect step is found.  \label{tab:error-finding-iterative}}
\end{table}

\begin{table}
    \centering
    \centering
    \resizebox{\linewidth}{!}{\begin{tabular}{lccc|c}
        Model & \multicolumn{3}{c}{{\color{mygreen}Overall Verification}} & {\color{mygreen}Stepwise}\\
         & Corr. F1 & Err. F1 & F1 & micro F1 \\
        \hline
        \multicolumn{5}{l}{\rule{0pt}{0.1in}\bf Few-shot} \\
        \;GPT3.5 & 0.75 & 0.65 & 0.70 & 0.50\\
        \;\; + solution & {\bf 0.91} & {\bf 0.88} & \textbf{0.89} & 0.63 \\
        \;Llama2-70B & 0.07  & 0.62  & 0.34 & 0.11 \\
        \;\; + solution & 0.80  & 0.82  & 0.81 & 0.59\\
        \hline
    \end{tabular}}
    \caption{Stepwise verification results on the small existing dataset from human evaluation of Roscoe~\citep{golovneva2023roscoe}. The Stepwise verification contains multi-class classification results. The results of the models are consistent with our dataset.  \label{tab:error-finding-roscoe-dataset}}
\end{table}

\begin{figure}
    \begin{center}
        \resizebox{1.0\linewidth}{!}{\includegraphics{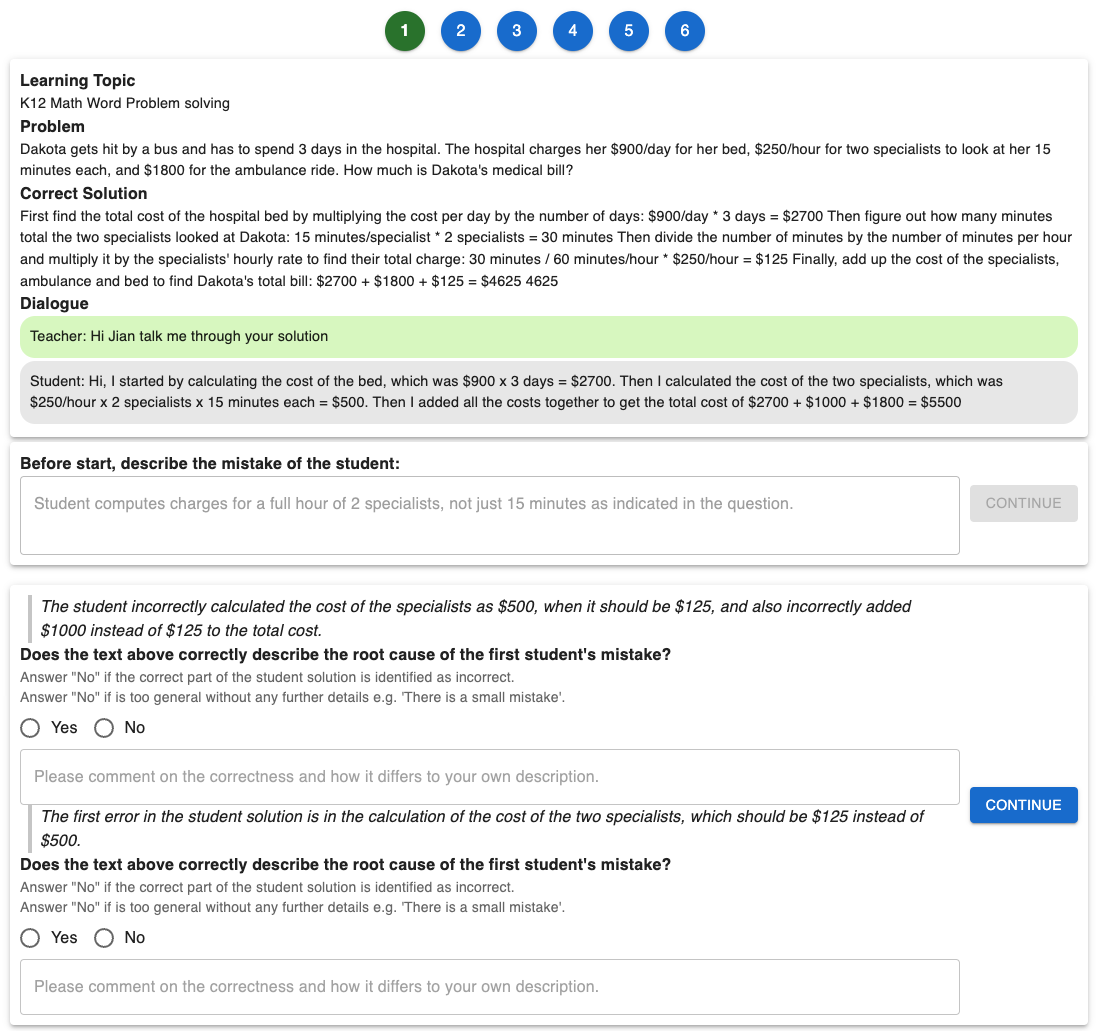}}
    \end{center}
    \caption{User interface for explaining the error of the student and evaluation of two error descriptions from models. Afterwards, annotators evaluate the quality of the model responses.
    \label{fig:evaluation-ui-error-description}}
\end{figure}

\begin{figure}
    \begin{center}
        \resizebox{1.0\linewidth}{!}{\includegraphics{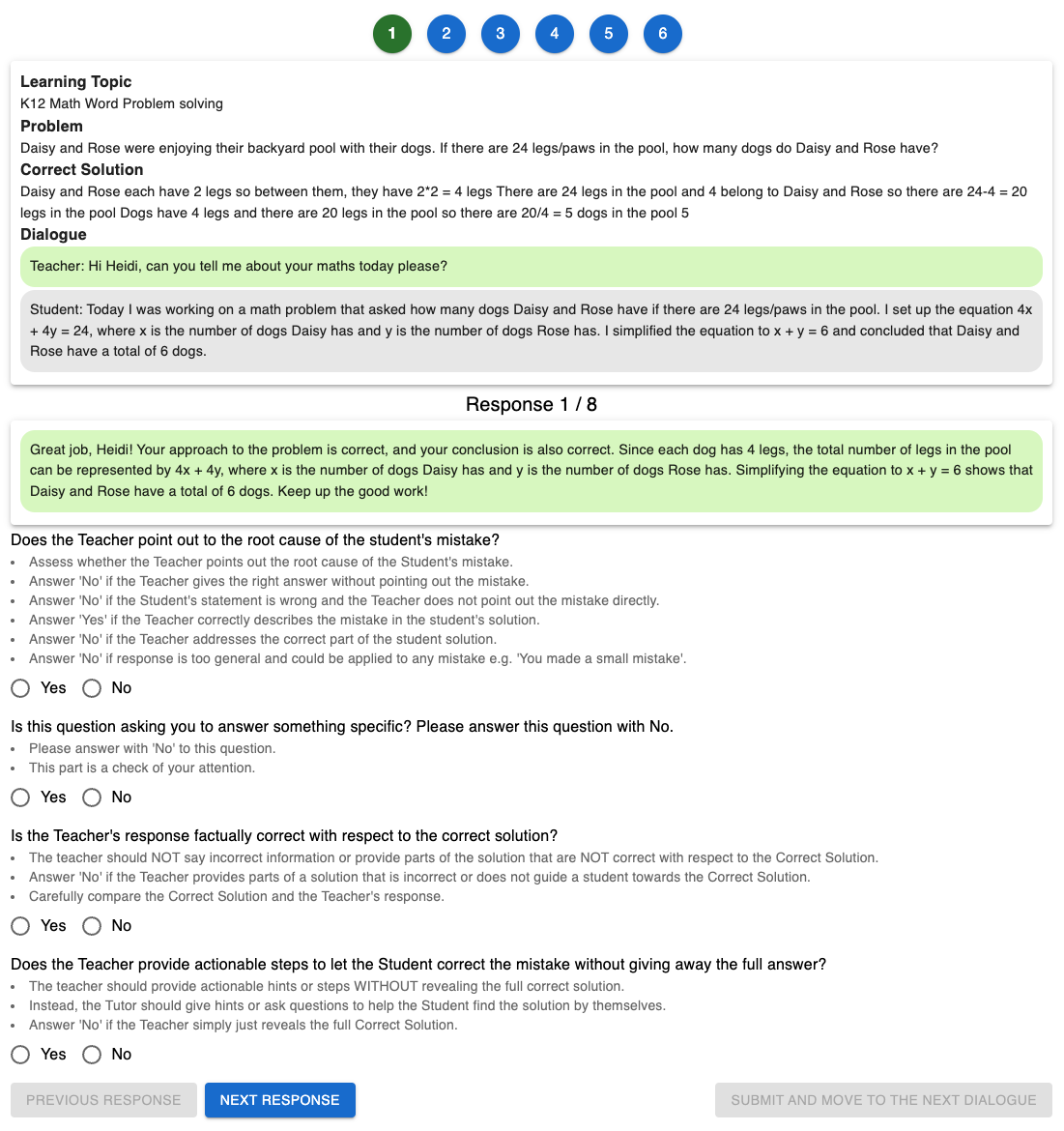}}
    \end{center}
    \caption{User interface for evaluation of the quality of the model responses. Some responses contain attention checks (second question in this case).
    \label{fig:evaluation-ui}}
\end{figure}

\section{Guidelines for Human Evaluation}\label{appendix-human-eval}
The user interface used in the human evaluation is shown in~\Cref{fig:evaluation-ui-error-description} and~\Cref{fig:evaluation-ui}. All the annotators had to complete a training for the task where each of their responses was evaluated and the feedback was provided to them. We used the subset of the annotators from~\Cref{appendix:step-errors-collection} with the same selection conditions and the same payment. Before evaluating the quality of responses, annotators are asked to analyze the math problem and the conversation and explain the student error in an open-ended text. To not bias their understanding of the student solution only subsequently the error descriptions from verifiers were annotated with their correctness using these instructions: \texttt{Does the text above correctly describe the root cause of the first student's mistake? Answer "No" if the correct part of the student solution is identified as incorrect. Answer "No" if is too general without any further details e.g. 'There is a small mistake'.}

The exact wording of the annotation questions for evaluating the quality of responses is the following: 

\paragraph{Targeted} \texttt{Does the Teacher point out to the root cause of the student's mistake? Answer 'No' if the Teacher gives the right answer without pointing out the mistake. Answer 'No' if the Student's statement is wrong and the Teacher does not point out the mistake directly. Answer 'Yes' if the Teacher correctly describes the mistake in the student's solution. Answer 'No' if the Teacher addresses the correct part of the student solution. Answer 'No' if response is too general and could be applied to any mistake e.g. 'You made a small mistake'.}

\paragraph{Correctness} \texttt{Is the Teacher's response factually correct with respect to the reference solution? The teacher should NOT say incorrect information or provide parts of the solution that are NOT correct with respect to the reference solution. Answer 'No' if the Teacher provides parts of a solution that is incorrect or does not guide a student towards the reference solution. Carefully compare the reference solution and the Teacher's response.}

\paragraph{Actionable} \texttt{Does the Teacher provide actionable steps to let the Student correct the mistake without giving away the full answer? The teacher should provide actionable hints or steps WITHOUT revealing the full reference solution. Instead, the Tutor should give hints or ask questions to help the Student find the solution by themselves. Answer 'No' if the Teacher simply just reveals the full reference solution.}

\section{\textcolor{myorange}{Alignment} Details}\label{alignment-appendix}
To find the best hyperparameters for the Alignment algorithm we run a grid search using values of similarity threshold $t=[0.5, 0.6, 0.7, 0.8, 0.9, 0.95]$ and gap costs $c=[-0.1, -0.2, -0.3, -0.5, -0.7, -1.0, -1.2]$. The best hyperparameters are reported in~\Cref{tab:alignment-eval}. The exact models which are used for semantic similarity are SBERT (\textit{sentence-transformers/all-mpnet-base-v2}) and Roscoe (\textit{facebook/roscoe-512-roberta-base}).

We use the template to transform the output of the algorithm into the textual prompt. In the  template, all the steps from the student solution and reference solution are used. Furthermore, the cost of the alignment can be used to filter out student solutions that differ completely from reference solution which we leave for future work. The template is the following:

\noindent\texttt{Missing steps in student solution: \{missing steps\}}\\
\noindent\texttt{Unnecessary steps in the student solution:  \{unnecessary steps\}}\\
\noindent\texttt{Matching steps: \{matching steps\}}

\section{Details on LLM-based Evaluation}
\label{app:llm-eval}
A response is targeted if it targets the students' mistake, correct if it does not conflict with grounding information, and actionable if it provides the student with useful guidance to help the student progress in their solution attempt.
In all cases, for each quality dimension, we provide the model with three examples (3-shots). We use the LLAMA3-70B\footnote{meta-llama/Meta-Llama-3-70B-Instruct} with temperature $T=0$ for reproducibility. The task description and the examples are the same as in the human evaluation for instructing the annotators described in~\Cref{sec:human_eval}. The prompt also includes the reference solution for more reliable judging~\citep{zheng2024judging,learnlm2024}.

\section{Qualitative examples}
\label{sec:qualitative_examples}
In this section, we show qualitative examples to better understand the behavior of verification and verification-based response generation.
We first show examples for prompted models in~\cref{tab:gpt-generated-examples} and then show examples for finetuned models in~\cref{tab:finetuned-ex-1}.

\newpage

\begin{table*}
\begin{tabularx}{\textwidth}{|X|}
\hline
\textbf{Problem:} At a cafe, the breakfast plate has two eggs and twice as many bacon strips as eggs.  If 14 customers order breakfast plates, how many bacon strips does the cook need to fry? \\
\hline
\textbf{Correct Solution:} Each breakfast plate has 2 x 2 = 4 bacon strips. The cook needs to fry 14 x 4 = 56 bacon strips.\\
\hline
\textbf{Error Description Verification:} The student incorrectly wrote the number of bacon strips per plate as 2 + 2x, where x is the number of eggs, \textcolor{mygreen}{instead of simply 2 x 2 = 4 bacon strips per plate}. This led to an incorrect total of 84 bacon strips. \\
\hline
\textbf{Dialogue} \\
\hline
\textit{Teacher:} Hi Claire could you please walk me through your solution \\
\textit{Student:} Sure! For each breakfast plate, there are 2 eggs and twice as many bacon strips as eggs. So, for each breakfast plate, there are 2 + 2x bacon strips, where x is the number of eggs. Since each breakfast plate has two eggs, x = 2. So, for each breakfast plate, there are 2 + 2x bacon strips = 2 + 2(2) = 6 bacon strips. Thus, for 14 breakfast plates, the cook needs to fry 6 x 14 = 84 bacon strips.\\
\hline
\textbf{Generated Responses}
\\
\hline
\textbf{Human Teacher:} Let's work on simpler problem. If I have 2 dogs for every cat and I have 5 cats, I should have double the number of dogs. How many dogs will I have \\
\hline
\textbf{Baseline:} It looks like you made a small mistake in your solution. You correctly found the number of bacon strips for one breakfast plate, but you used the wrong value for x. \textcolor{red}{Since each breakfast plate has 2 eggs, x should be 2, not 2 eggs. So the correct number of bacon strips for one breakfast plate is 2 + 2(2) = 6, and for 14 plates, the cook needs to fry 6 x 14 = 84 bacon strips.} Great effort! \\
\hline
\textbf{Error reason:} It looks like you made a small mistake in your calculation. You correctly found that for each breakfast plate, there are 6 bacon strips, but when you multiplied by 14, \textcolor{red}{you should have gotten 84, not 6 x 14 = 84. So the correct answer is 6 bacon strips per plate multiplied by 14 plates equals 84 bacon strips.} \\
\hline
\textbf{Error description:} It looks like there was a mistake in your calculation for the number of bacon strips for each breakfast plate. \textcolor{mygreen}{Remember that the breakfast plate has twice as many bacon strips as eggs, so for each plate, there are 2 eggs and 4 bacon strips.} Can you try recalculating the total number of bacon strips needed for 14 breakfast plates? \\
\hline
\textbf{Alignment:} It looks like you made a small mistake in your solution. \textcolor{mygreen}{The breakfast plate has 2 eggs, so it has 2 x 2 = 4 bacon strips. If 14 customers order breakfast plates, the cook needs to fry 14 x 4 = 56 bacon strips.} The answer is 56. \\
\hline
\end{tabularx}
\caption{Examples of responses generated by GPT-3.5 prompted models for the same problem.}
\label{tab:gpt-generated-examples}
\end{table*}

\begin{table*}
\begin{tabularx}{\textwidth}{|X|}
\hline
\textbf{Problem:} A pet store owner had 12 birds, 9 puppies, 5 cats, and 15 spiders. Half the birds were sold and 3 puppies were adopted. Then, someone left the spider enclosure open and 7 of them went loose. How many animals does the pet store owner have left?  \\
\hline
\textbf{Correct Solution:} There are 12/2 = 6 birds left. There are 9-3= 6 puppies left. There are 15 - 7 = 8 spiders left. The pet store owner has 5+6 + 6 + 8 = 25 animals left.\\
\hline
\textbf{Verification:} The student {\color{mygreen} failed to include the 5 cats} in their final count, resulting in an incorrect total of 20 animals instead of 25. \\
\hline
\textbf{Dialogue} \\
\hline
\textit{Teacher:} Hi , could you please walk me through your solution? \\
\textit{Student:} Sure, I started by taking half of the 12 birds, which is 6 birds. Then, I subtracted 6 from 12 to get the number of birds left, which is 6. After that, I subtracted 3 from 9 to get the number of puppies left, which is 6. Finally, I subtracted 7 from 15 to get the number of spiders left, which is 8. So, the pet store owner has 6 birds, 6 puppies, and 8 spiders left, which adds up to 20 animals.  \\
\hline
\textbf{Generated Responses} \\ \hline
\textit{Human Teacher:} i think {\color{mygreen} you missed something. Cats?} \\
\textit{Baseline:} How many birds were sold and how many puppies were adopted? \\
\textit{Ours:} {\color{mygreen} How many cats did the pet store owner have?} \\
\hline
\hline
\textbf{Problem:} Every year, Tabitha adds a new color to her hair.  She started this tradition when she was 15 years old, which was the year she added her second hair color.  In three years, Tabitha will have 8 different colors in the hair.  Currently, how old is Tabitha?  \\
\hline
\textbf{Correct solution:} {\color{mygreen} Since, three years from now, Tabitha will have 8 different colors in her hair, this year she has 8 - 3 = 5 colors in her hair.} Since she had 2 colors in her hair at the age of 15, she has added 5-2=3 colors since she was 15 years old.
Since she ads one color per year, 3 added colors = 3 added years. Thus, currently, she is 15+3=18 years old.
\\
\hline
\textbf{Verification:} The student incorrectly assumed that Tabitha is currently 21 years old, without considering the information that in three years, she will have 8 different colors in her hair. The student failed to work backwards to find Tabitha's current age.  \\
\hline
\textbf{Dialogue} \\
\hline
\textit{Teacher:} Hi Jordy. Tell me about your answer \\
\textit{Student:} I think my answer is correct. I used a calculator and I didn't make any calculation errors. I figured that since Tabitha started adding a new color to her hair when she was 15, and she has added a new color every year since then, she must be 21 years old now.  \\
\hline
\textbf{Generated Response} \\ \hline
\textit{Human:} Let's approach this working backward. If in three years, Tabitha will have 8 colors in her hair, {\color{mygreen} how many colors does she have in her hair now?}  \\
\textit{Baseline:} How many years has Tabitha been adding a new color to her hair? \\
\textit{Ours:} {\color{mygreen} How many different colors does Tabitha have in her hair now?}  \\
\hline
\end{tabularx}
\caption{Qualitative examples of finetuned response generation models.}
\label{tab:finetuned-ex-1}
\end{table*}

\section{Prompts}
\label{app:prompts}
This section provides the exact prompts used in our work.
First, we show the prompt used for the baseline, error description-based, and alignment-based response generation models in~\cref{fig:prompt-baseline} and \cref{fig:prompt-description-and-alignment-generation}. Verification prompts for Error Description are in~\cref{fig:prompt-description-intermediate} and for Error Reason in~\cref{fig:prompt-error-reason-verification}. The prompt with 5 examples for the CoT solution generation is in~\cref{fig:prompt-cot-solution}. 
Then, we show the prompts used for targeted LLM-based evaluation in \cref{fig:prompt-critic-targeted}, correctness evaluation in \cref{fig:prompt-critic-correctness}, and evaluation of how actionable responses are in \cref{fig:prompt-critic-actionable}. To sample responses from models by prompting we use temperature $T=0$ for reproducibility. 

\begin{figure*}[ht]
    \centering
    \small
    \begin{tcolorbox}
        You are an experienced teacher and you are going to respond to a student. The problem your student is solving is on topic: \{topic\}. \\
        Problem: \{problem\} \\
        \{conversation\} \\
        Teacher (maximum two sentences):
    \end{tcolorbox}
    \caption{
    Response generation prompt for the \textbf{direct baseline}.
    \texttt{\{problem\}} is a placeholder for the problem the student is solving,
    \texttt{\{topic\}} is the learning topic, and 
    \texttt{\{conversation\}} is a conversation history. 
    \label{fig:prompt-baseline}}
\end{figure*}

\begin{figure*}[ht]
    \centering
    \small
    \begin{tcolorbox}
        You are an experienced teacher. Your task is to read a conversation snippet of a tutoring session between a student and tutor, and determine what type of error the student makes in the conversation. We have a list of common errors that students make in math, which you can pick from. We also give you the option to write in your own error type if none of the options apply.\\
        Error list:\\
        0. Student does not seem to understand or guessed the answer.\\
        1. Student misinterpreted the question.\\
        2. Student made a careless mistake.\\
        3. Student has the right idea, but is not quite there.\\
        4. Student’s answer is not precise enough or the tutor is being too picky about the form
        of the student’s answer.\\
        5. None of the above, but I have a different description (please specify in your
        reasoning).\\
        6. Not sure, but I’m going to try to diagnose the student. \\
        Here is the conversation snippet:\\
        Lesson topic: \{topic\}.\\
        Problem: \{problem\} \\
        \{conversation\} \\
        Why do you think the student made this mistake? Pick an option number from the error list and provide the reason behind your choice. Format your answer as: \{"answer": \#, "reason": "write out your reason for picking \# here"\} \\
    \end{tcolorbox}
    \caption{
    Verification for \textbf{Error reason baseline}~\citep{wang-etal-2024-bridging}.
    \texttt{\{topic\}} is the learning topic, \texttt{\{problem\}} is a placeholder for the problem the student is solving, and 
    \texttt{\{conversation\}} is a conversation history. 
    \label{fig:prompt-error-reason-verification}}
\end{figure*}

\begin{figure*}[ht]
    \centering
    \small
    \begin{tcolorbox}
        You are an experienced math teacher. Your goal is to identify the correctness of the Student's Solution to a Problem. \\
        Problem: \{problem\} \\
        Expected reference solution: \{solution\} \\
        \{conversation\} \\
        Q: Find the first error in the student solution compared to the expected reference solution and write a one line description. If no error, write "Student' solution is Correct". \\
        A:
    \end{tcolorbox}
    \caption{
    Verification prompt for \textcolor{myblue}{Error description} of the first student error.
    \texttt{\{problem\}} is a placeholder for the problem the student is solving,
    \texttt{\{solution\}} is a solution generated from the same model using CoT prompt in~\Cref{fig:prompt-cot-solution}, and 
    \texttt{\{conversation\}} is a conversation history. 
    \label{fig:prompt-description-intermediate}}
\end{figure*}

\begin{figure*}[ht]
    \centering
    \small
    \begin{tcolorbox}
        You are an experienced teacher and you are going to respond to a student. The problem your student is solving is on topic: \{topic\}. \\
        Problem: \{problem\} \\
        Assessment of student solution: \{description\} \\
        \{conversation\} \\
        Teacher (maximum two sentences): 
    \end{tcolorbox}
    \caption{
    Response generation for \textbf{Error reason baseline}, \textcolor{myblue}{Error description}, and \textcolor{myorange}{Alignment generation}.
    \texttt{\{problem\}} is a placeholder for the problem the student is solving,
    \texttt{\{topic\}} is the learning topic,
    \texttt{\{conversation\}} is a conversation history, \texttt{\{description\}} is the result of the particular verification step.}
    \label{fig:prompt-description-and-alignment-generation}
\end{figure*}

\begin{figure*}[ht]
    \centering
    \small
    \begin{tcolorbox}
        You are a highly intelligent question answering assistant. Solve the question step-by-step. Always finish the answer by providing your final answer after 'The answer is'. \\
        Question: Natalia sold clips to 48 of her friends in April, and then she sold half as many clips in May. How many clips did Natalia sell altogether in April and May? \\
        Answer: Natalia sold 48/2 = <<48/2=24>>24 clips in May. Natalia sold 48+24 = <<48+24=72>>72 clips altogether in April and May. The answer is 72 \\
        Question: Julie is reading a 120-page book. Yesterday, she was able to read 12 pages and today, she read twice as many pages as yesterday. If she wants to read half of the remaining pages tomorrow, how many pages should she read? \\
        Answer: Maila read 12 x 2 = <<12*2=24>>24 pages today. So she was able to read a total of 12 + 24 = <<12+24=36>>36 pages since yesterday. There are 120 - 36 = <<120-36=84>>84 pages left to be read. Since she wants to read half of the remaining pages tomorrow, then she should read 84/2 = <<84/2=42>>42 pages. The answer is 42 \\
        Question: Weng earns \$12 an hour for babysitting. Yesterday, she just did 50 minutes of babysitting. How much did she earn? \\
        Answer: Weng earns 12/60 = <<12/60=0.2>>0.2 per minute. Working 50 minutes, she earned 0.2 x 50 = <<0.2*50=10>>10. The answer is 10 \\
        Question: The profit from a business transaction is shared among 2 business partners, Mike and Johnson in the ratio 2:5 respectively. If Johnson got \$2500, how much will Mike have after spending some of his share on a shirt that costs \$200? \\
        Answer: According to the ratio, for every 5 parts that Johnson gets, Mike gets 2 parts Since Johnson got \$2500, each part is therefore \$2500/5 = \$<<2500/5=500>>500 Mike will get 2*\$500 = \$<<2*500=1000>>1000. After buying the shirt he will have \$1000-\$200 = \$<<1000-200=800>>800 left. The answer is 800 \\
        Question: Ralph is going to practice playing tennis with a tennis ball machine that shoots out tennis balls for Ralph to hit. He loads up the machine with 175 tennis balls to start with. Out of the first 100 balls, he manages to hit 2/5 of them. Of the next 75 tennis balls, he manages to hit 1/3 of them. Out of all the tennis balls, how many did Ralph not hit? \\
        Answer: Out of the first 100 balls, Ralph was able to hit 2/5 of them and not able to hit 3/5 of them, 3/5 x 100 = 60 tennis balls Ralph didn't hit. Out of the next 75 balls, Ralph was able to hit 1/3 of them and not able to hit 2/3 of them, 2/3 x 75 = 50 tennis balls that Ralph didn't hit. Combined, Ralph was not able to hit 60 + 50 = <<60+50=110>>110 tennis balls Ralph didn't hit. The answer is 110 \\
        Question: \{problem\} \\
        Answer:
    \end{tcolorbox}
    \caption{
    Prompt for the chain-of-thought (CoT) reference solution generation.
    \texttt{\{problem\}} is a placeholder for the problem the student is solving.
    \label{fig:prompt-cot-solution}}
\end{figure*}

\begin{figure*}[ht]
    \small
    \begin{tcolorbox}
        You are a Critic giving feedback on the Teacher's tutoring skills. As the Critic you assess whether the Teacher points out the root cause of the Student's mistake. \\ 
        Answer 'No' if the Student's statement is wrong and the Teacher does not point out the mistake directly. Answer 'No' if the Teacher gives the right answer without pointing out the mistake. Answer 'Yes' if the Teacher correctly describes the mistake in the student's solution. Answer 'No' if the Teacher addresses the correct part of the student solution. Answer 'No' if response is too general and could be applied to any mistake e.g. 'You made a small mistake'. \\
        Does the Teacher points out to the root cause of the student's mistake?  Output a 'Yes' or 'No' decision and a short rationale in the format Decision: <decision>. Rationale: <rationale>. \\\\
        \lbrack Examples\rbrack \\
        Problem: The profit from a business transaction is shared among 2 business partners, Mike and Johnson in the ratio 2:5 respectively. If Johnson got \$2500, how much will Mike have after spending some of his share on a shirt that costs \$200? \\
        Solution: According to the ratio, for every 5 parts that Johnson gets, Mike gets 2 parts. Since Johnson got \$2500, each part is therefore \$2500/5 = \$500. Mike will get 2*\$500 = \$1000. After buying the shirt he will have \$1000-\$200 = \$800 left
        \lbrack Dialogue\rbrack \\
        Teacher: Could you please walk me through your solution? \\
        Student: Sure. I figured that Jonsons' profit of \$2500 is equivalent to 7 parts, so each part is worth \$357.14. Since Mike got 2 parts, his total portion is 714.28, of which he spent 200, leaving him with 514.28\\
        \lbrack Example 1\rbrack \\
        Teacher response: Great job on correctly calculating the value of each part and determining Mike's remaining amount after spending on the shirt. Your final answer is correct. \\
        Request: Does the Teacher point out the root cause of the student's mistake?  \\
        Critic: Decision: No. Rationale: The teacher does not identify any error in the student's solution. Moreover, it acknowledges the student's incorrect solution as correct.\\
        \lbrack Example 2\rbrack \\
        Teacher response: Your calculation is incorrect. Can you try to double-check the parts each gets? \\
        Request: Does the Teacher point out the root cause of the student's mistake? \\
        Critic: Decision: No. Rationale: The teacher just generally state your solution is incorrect and this is not specific to the student error. \\
        \lbrack Example 3\rbrack \\ 
        Teacher response: Johnson's \$2500 represents 5 parts, not 7, so each part is \$500, and Mike's share before buying the shirt is \$1000, not \$714.28. So the solution is \$800. \\
        Request: Does the Teacher point out the root cause of the student's mistake? \\
        Critic: Decision: Yes. Rationale: The Teacher provides very specific identification by directly stating the problem is in using the wrong number of parts. \\
        \lbrack The End of Examples\rbrack\\\\
        Problem: \{problem\} \\ 
        Reference solution: \{correct answer\} \\
        \lbrack Dialogue\rbrack \\ 
        \{dialog history\} \\
        Teacher response: \{response\} \\ 
        Request: Does the Teacher point out the root cause of the student's mistake? \\
        Critic:
    \end{tcolorbox}
    \caption{Prompt for targeted evaluation.\label{fig:prompt-critic-targeted}}
\end{figure*}

\begin{figure*}[ht]
    \small
    \begin{tcolorbox}
        You are a Critic giving feedback on the correctness of the Teacher who is interacting with a Student.  \\
        The teacher should NOT say incorrect information or provide parts of the solution that are NOT correct with respect to the reference solution. 
        Answer 'No' if the Teacher provides parts of a solution that is incorrect or does not guide a student towards the reference solution.\\
        Is the Teacher's response factually correct with respect to the reference solution? Output a 'Yes' or 'No' decision and a short rationale in the format Decision: <decision>. Rationale: <rationale>. \\
        Carefully compare the reference solution and the Teacher's response. \\\\
        \lbrack Examples\rbrack \\
        Problem: The profit from a business transaction is shared among 2 business partners, Mike and Johnson in the ratio 2:5 respectively. If Johnson got \$2500, how much will Mike have after spending some of his share on a shirt that costs \$200? \\
        Solution: According to the ratio, for every 5 parts that Johnson gets, Mike gets 2 parts. Since Johnson got \$2500, each part is therefore \$2500/5 = \$500. Mike will get 2*\$500 = \$1000. After buying the shirt he will have \$1000-\$200 = \$800 left \\
        \lbrack Dialogue\rbrack \\
        Teacher: Could you please walk me through your solution? \\
        Student: Sure. I figured that Jonsons' profit of \$2500 is equivalent to 7 parts, so each part is worth \$357.14. Since Mike got 2 parts, his total portion is 714.28, of which he spent 200, leaving him with 514.28 \\
        \lbrack Example 1\rbrack \\
        Teacher response: Your calculation is incorrect. Can you try to double-check the parts each gets? \\
        Request: Is the Teacher's response factually correct with respect to the reference solution? \\
        Critic: Decision: Yes. Rationale: The Teacher's response correctly states there is a mistake in the student's calculation and ask a question. Nothing factually incorrect is said. \\
        \lbrack Example 2\rbrack \\ 
        Teacher response: Johnson's \$2500 represents 5 parts, not 7, so each part is \$500, and Mike's share before buying the shirt is \$1000, not \$714.28. So the solution is \$800. \\
        Request: Is the Teacher's response factually correct with respect to the reference solution? \\
        Critic: Decision: Yes. Rationale: The Teacher's response is stating part of the reference solution and it is factually correct. \\
        \lbrack Example 3\rbrack \\
        Teacher response: Great job on correctly calculating the value of each part and determining Mike's remaining amount after spending on the shirt. Your final answer is correct. \\
        Request: Is the Teacher's response factually correct with respect to the reference solution? \\ 
        Critic: Decision: No. Rationale: The Teacher acknowledges the student's incorrect solution as correct, which is not factually correct given the correct answer. \\
        \lbrack The End of Examples\rbrack \\\\
        Problem: \{problem\} \\ 
        Reference solution: \{correct answer\} \\
        \lbrack Dialogue\rbrack \\ 
        \{dialog history\} \\
        Teacher response: \{response\} \\ 
        Request: Is the Teacher's response factually correct with respect to the reference solution? \\
        Critic:
    \end{tcolorbox}
    \caption{Prompt for correctness evaluation.\label{fig:prompt-critic-correctness}}
\end{figure*}

\begin{figure*}[ht]
    \small
    \begin{tcolorbox}
        You are a Critic giving feedback on the responses of the Teacher who is interacting with a Student. \\
        Your task is to gauge if the Teacher's Response provides actionable hints or steps without revealing the full solution.
        The Student could use this response to move closer to the final correct answer. 
        A good response could also be a follow-up question that makes the user think about how to solve the problem. \\
        Does the Teacher provide actionable steps without giving away the full answer? Output a 'Yes' or 'No' decision and a short rationale in the format Decision: <decision>. Rationale: <rationale>. \\
        Answer 'No' if the Teacher simply just reveals the full reference solution. \\\\
        \lbrack Examples\rbrack \\
        Problem: The profit from a business transaction is shared among 2 business partners, Mike and Johnson in the ratio 2:5 respectively. If Johnson got \$2500, how much will Mike have after spending some of his share on a shirt that costs \$200? \\
        Solution: According to the ratio, for every 5 parts that Johnson gets, Mike gets 2 parts. Since Johnson got \$2500, each part is therefore \$2500/5 = \$500. Mike will get 2*\$500 = \$1000. After buying the shirt he will have \$1000-\$200 = \$800 left \\
        \lbrack Dialogue\rbrack \\
        Teacher: Could you please walk me through your solution? \\
        Student: Sure. I figured that Jonsons' profit of \$2500 is equivalent to 7 parts, so each part is worth \$357.14. Since Mike got 2 parts, his total portion is 714.28, of which he spent 200, leaving him with 514.28 \\  
        \lbrack Example 1\rbrack \\
        Teacher response: Your calculation is incorrect. Can you try to double-check the parts each gets? \\
        Request: Does the Teacher provide actionable steps without giving away the full answer? \\
        Critic: Decision: Yes. Rationale: the teacher asks a question or ask for action from the student to double-check the answer. \\
        \lbrack Example 2\rbrack \\
        Teacher response: Johnson's \$2500 represents 5 parts, not 7, so each part is \$500, and Mike's share before buying the shirt is \$1000, not \$714.28. So the solution is \$800. \\
        Request: Does the Teacher provide actionable steps without giving away the full answer? \\
        Critic: Decision: No. Rationale: The teacher states the reference solution at the end. \\
        \lbrack Example 3\rbrack \\
        Teacher response: Johnson's \$2500 represents 5 parts, not 7, so each part is \$500, and Mike's share before buying the shirt is \$1000, not \$714.28. \\ 
        Request: Does the Teacher provide actionable steps without giving away the full answer? \\
        Critic: Decision: Yes. Rationale: The teacher points out what is wrong with the student's solution but do not tell the full correct answer. \\
        \lbrack The End of Examples\rbrack \\\\
        Problem: \{problem\} \\ 
        Reference solution: \{correct answer\} \\
        \lbrack Dialogue\rbrack \\ 
        \{dialog history\} \\
        Teacher response: \{response\} \\ 
        Request: Does the Teacher provide actionable steps without giving away the full answer? \\
        Critic:
    \end{tcolorbox}
    \caption{Prompt for actionable evaluation.\label{fig:prompt-critic-actionable}}
\end{figure*}

\section{Finetuning Details}\label{finetuning-appendix}
We finetune all models by extending the huggingface transformers library~\citep{wolf-etal-2020-transformers} and using the checkpoints from the huggingface hub in accordance with the corresponding license agreements.

For verification, we finetune LLAMA2 with 7B parameters and using LoRA.
We use a learning rate of $1\cdot10^{-5}$, linear learning rate decay with 32 warmup steps, a batch size of 2 and train for 6 epochs in total.

For response generation, we finetune Flan-T5 3B with LoRA with a learning rate of $1\cdot10^{-5}$, a batch size of 2 and a total of 10 training epochs.

For both tasks, we used NVIDIA A100 80GB GPU and training takes around 3-6 hours for 5 or 10-fold cross-validation.

\end{document}